# Projection-wise Disentangling for Fair and Interpretable Representation Learning: Application to 3D Facial Shape Analysis

Xianjing Liu[1], Bo Li[1], Esther Bron[1], Wiro Niessen[1,2], Eppo Wolvius[1] and Gennady Roshchupkin[1]

[1] Erasmus MC, Rotterdam, The Netherlands
[2] Delft University of Technology, Delft, The Netherlands
`g.roshchupkin@erasmusmc.nl`

**Abstract.** Confounding bias is a crucial problem when applying machine learning to practice, especially in clinical practice. We consider the problem of learning representations independent to multiple biases. In literature, this is mostly solved by purging the bias information from learned representations. We however expect this strategy to harm the diversity of information in the representation, and thus limiting its prospective usage (e.g., interpretation). Therefore, we propose to mitigate the bias while keeping almost all information in the latent representations, which enables us to observe and interpret them as well. To achieve this, we project latent features onto a learned vector direction, and enforce the independence between biases and projected features rather than all learned features. To interpret the mapping between projected features and input data, we propose projection-wise disentangling: a sampling and reconstruction along the learned vector direction. The proposed method was evaluated on the analysis of 3D facial shape and patient characteristics (N=5011). Experiments showed that this conceptually simple method achieved state-of-the-art fair prediction performance and interpretability, showing its great potential for clinical applications.

**Keywords:** Fair representation learning, Disentangled representation learning, Interpretability, 3D shape analysis.

## 1 Introduction

Machine learning techniques, especially deep learning, have emerged as a powerful tool in many domains. However, its susceptibility to bias present in training datasets and tasks poses, brings a new challenge for the practical applicability, i.e., spurious performance in the training and evaluation stage with limited generalizability to application in new conditions [1]. To mitigate (confounding) bias in data analysis, traditional statistic methods use special techniques such as control-matching [2] and stratification [3]. However, due to the end-to-end training scheme and the need for large-size training data, these techniques are no longer favored by the machine learning field.

In "*Representation Learning*" one tries to find representations (i.e., learned features, $\mathbf{Z}$) of the data that are related to specific attributes (i.e., the learning target, $t$).

Especially, a *fair representation* means it contains no information of sensitive attributes (i.e., bias, $s$) [8]. Existing methods for fair representation learning can be categorized into two types: 1) adversarial training, in which methods are trained to predict the bias from the representation, and subsequently minimize the performance of the adversary to remove bias information from the representation [4,5,6,7], and 2) variational auto-encoder (VAE) -based methods [8,9,10,11], which minimize the dependency between the latent representation and the bias using Mutual Information (MI) or Maximum Mean Discrepancy (MMD) metrics [8,10]. Despite their potential to facilitate fair representation, these models are not interpretable, which can limit their applicability in clinical practice [12,13]. Moreover, the fairness of these methods is approached by purging all bias information from the learned representations (i.e., $MI(\mathbf{Z}, s) \to 0$). This strict strategy can reduce the diversity of information in $\mathbf{Z}$ (Fig. A3).

To address these issues, we propose a novel projection-wise disentangling strategy for auto-encoder-based fair and interpretable representation learning. We construct $z_p$ as a linear projection of latent features onto a vector direction, and learn the fair representation by minimizing the correlation between $z_p$ and $s$, i.e., $MI(z_p, s) \to 0$. Compared with existing strategies of global-constraint $MI(\mathbf{Z}, s) \to 0$, the proposed conditional-constraint strategy can maintain the diversity of information in $\mathbf{Z}$ , and thus 1) obtains an **optimal trade-off** between reconstruction quality and fairness, fitting the proposed method into **semi-supervised extensions**; Also, 2) we propose projection-wise disentangling to **interpret** the disentanglement of correlated attributes; 3) Our method can easily handle **multiple and continuous biases**.

In this paper, we applied the proposed method to clinical applications of 3D facial shape analysis. For epidemiological studies the bias (confounding) problem is crucial, because it can create association that is not true, or association that is true, but misleading, thus leading to wrong diagnosis or therapy strategy. The bias problem becomes significant for AI since it has been used more and more for medical data analysis. While we applied our framework to 3D facial images due to our clinical interest, the approach can be generally used for other type of data.

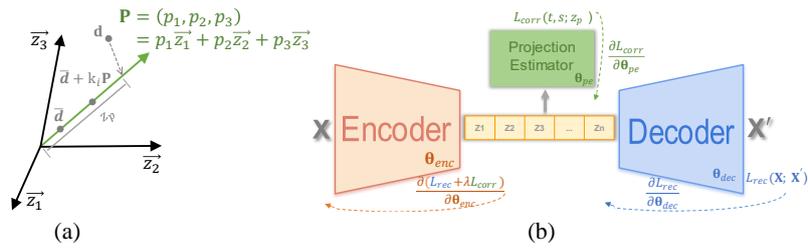

**Fig. 1.** a) $\mathbf{P} = [p_1, p_2, ... p_n]$ represents a vector in latent space ($n = 3$). $\mathbf{d}$ is the latent representation of a datapoint. $\mathbf{d}$ and $\mathbf{d} + k_i \mathbf{P}$ show the sampling along $\mathbf{P}$ in Eq. 6. When sampling along $\mathbf{P}$, $z_p$ changes and this change is correlated to $t$ ($|Corr(\mathbf{z_p}, \mathbf{t})|$ is maximized) while independent to $s$ ($|Corr(\mathbf{z_p}, \mathbf{s})|$ is minimized); b) Framework of the proposed method.

## 2 Methods

### 2.1 Projecting features onto a Direction in the Latent Space

A latent space can be viewed as a vector subspace of $\mathbb{R}^n$ ($n$ latent features) with basis vectors[1] $\vec{z}_i$ ($i = 1, 2 \ldots n$). Some recent studies show that most attributes (target or bias) of the input data have a vector direction that predominantly captures its variability [22,23]. We aim to couple a vector direction with the target while being independent to the bias. A vector direction in the latent space can be represented by $\mathbf{P} = [p_1, p_2, \ldots, p_n]$ as a linear combination of the basis vectors (Fig. 1a). Let $\mathbf{D^{d \times n}} = [\mathbf{z_1^d}, \mathbf{z_2^d}, \ldots, \mathbf{z_n^d}]$ ($d$ datapoints) be the latent representations of the input data. For each datapoint, $\mathbf{d} = [z_1, z_2, \ldots, z_n]$ is its latent representations, and $z_p = \mathbf{d} * \mathbf{P} / \|\mathbf{P}\| = (p_1 z_1 + p_2 z_2 + \cdots + p_n z_n) / \|\mathbf{P}\|$ can be viewed as a scalar projection of $\mathbf{d}$ onto vector $\mathbf{P}$ (Fig 1a).

The canonical correlation analysis theory suggests that a linear combination (projection) of multiple variables can have a maximum or minimum linear relationship with specific attributes [14]. We thus formulate the linear relationship between $\mathbf{z_p}$ and attributes as follow:

$$\mathbf{z_p} = \beta_0 + \beta_s \mathbf{s} + \beta_t \mathbf{t} + \varepsilon , \tag{1}$$

where $\mathbf{z_p}^\mathbf{d}$ is the projections of d datapoints, $\mathbf{s^d}$ denotes a bias and $\mathbf{t^d}$ denotes a target with d datapoints, $\varepsilon$ is the error term, $\beta_0$ is a constant, $\beta_s$ and $\beta_t$ denote the coefficients of $\mathbf{s}$ and $\mathbf{t}$.

Subsequently, we link a vector direction $\mathbf{P}$ with the target and bias, with the goal to estimate a $\mathbf{P}$ that minimizes $|\beta_s|$ (equivalent to minimizing $|Corr(\mathbf{z_p}, \mathbf{s})|$) while maximizing $|\beta_t|$ (equivalent to maximizing $|Corr(\mathbf{z_p}, \mathbf{t})|$). Here $Corr(.,.)$ is the Pearson correlation coefficient which ranges between $[-1, 1]$.

To promote a linear relation between $\mathbf{z_p}$ and $(\mathbf{t}, \mathbf{s})$, which is required for effective bias minimization (Eq. 1), a correlation loss (Eq. 3) is further proposed. The correlation loss not only estimates $\mathbf{P}$, but also encourages the encoder to generate $\mathbf{z_p}$ linearly correlated to $\mathbf{t}$ and $\mathbf{s}$.

### 2.2 Implementation in the Auto-encoder

Our projection-wise disentangling method is shown in Fig. 1b. Let $\mathbf{X}$ be the input data, $\mathbf{X}'$ be the reconstructed data; **ENC** be the encoder, **DEC** the decoder, and **PE** the projection estimator, parameterized by trainable parameters $\boldsymbol{\theta}_{enc}$, $\boldsymbol{\theta}_{dec}$, and $\boldsymbol{\theta}_{pe}$, i.e.: $\mathbf{Z} = \mathbf{ENC}(\mathbf{X};\ \boldsymbol{\theta}_{enc})$; $\mathbf{X}' = \mathbf{DEC}(\mathbf{Z};\ \boldsymbol{\theta}_{dec})$; $\mathbf{z_p} = \mathbf{PE}(\mathbf{Z};\ \boldsymbol{\theta}_{pe})$; $\boldsymbol{\theta}_{pe} = \mathbf{P} = [p_1, p_2, \ldots, p_n]$.
To extract latent features $\mathbf{Z}$, $\boldsymbol{\theta}_{enc}$ and $\boldsymbol{\theta}_{dec}$ can be optimized in an unsupervised way using a reconstruction loss ($L_{rec}$), as quantified by the mean squared error between the input data and the reconstructed data:

$$L_{rec}(\mathbf{X};\ \mathbf{X}') = \|\mathbf{X} - \mathbf{X}'\|^2 . \tag{2}$$

---

[1] Non-orthogonal basis vectors.

Based on Eq. 1, to estimate $\boldsymbol{\theta}_{pe}$, as well as to optimize $\boldsymbol{\theta}_{enc}$ and thereby promote the linear correlation between $\mathbf{z_p}$ and $(\mathbf{s}, \mathbf{t})$, we propose a correlation loss ($L_{corr}$):

$$L_{corr}(\mathbf{t}, \mathbf{s}; \mathbf{z_p}) = |Corr(\mathbf{z_p}, \mathbf{s})| - \eta |Corr(\mathbf{z_p}, \mathbf{t})|, \tag{3}$$

where $\eta$ can be considered as a Lagrange multiplier to balance the correlations. $L_{corr}$ can handle binary and continuous biases. In case $s$ is categorical, it can be converted into dummy variables. Besides, it can be easily extended to handle tasks with multiple biases $\mathbf{s_i^d}$ ($i = 1, 2, \ldots m$):

$$L_{corr}(\mathbf{t}, \mathbf{s_1}, \mathbf{s_2}, \ldots, \mathbf{s_m}; \mathbf{z_p}) = |Corr(\mathbf{z_p}, \mathbf{s_1})| + \cdots + |Corr(\mathbf{z_p}, \mathbf{s_m})| - \eta |Corr(\mathbf{z_p}, \mathbf{t})|. \tag{4}$$

Combining Eq. 2 and 3, we optimize the proposed framework using a multi-task loss function ($L_{joint}$):

$$\boldsymbol{\theta}_{enc}, \boldsymbol{\theta}_{dec}, \boldsymbol{\theta}_{pe} \leftarrow argmin \ L_{joint} = L_{rec} + \lambda L_{corr}, \tag{5}$$

where $\lambda$ balances the magnitude of the reconstruction quality and the fairness terms.

In addition, for applications where (target and bias) attributes are only available for part of the data, we provide a semi-supervised implementation of the method to fully exploit the data. In particular, for each training batch, we update the parameters in two steps: 1) update by $L_{rec}$ based on the unlabelled data (half batch), and 2) update by $L_{joint}$ based on the labelled data (half batch). A detailed implementation is provided in the supplementary file (Algorithm 1).

### 2.3 Projection-wise Disentangling for Interpretation

When $\boldsymbol{\theta}_{enc}$, $\boldsymbol{\theta}_{dec}$ and $\boldsymbol{\theta}_{pe}$ are determined, for each given input, its fair projection $z_p$ can be estimated and used for fair prediction of $t$ with logistic or linear regression (LR), i.e., $\hat{t} = LR(z_p)$. To provide insight into the effect of fair representation on feature extraction, we visualize the reconstructed images and establish its correspondence with the target attribute $\hat{t}_i$:

$$\mathbf{X}_i' = \mathbf{DEC}(\bar{\mathbf{d}} + \mathbf{k}_i \mathbf{P}; \ \boldsymbol{\theta}_{dec}) \text{ , and}$$

$$\hat{t}_i = LR\big(\mathbf{PE}(\bar{\mathbf{d}} + \mathbf{k}_i \mathbf{P}; \ \boldsymbol{\theta}_{pe})\big), \ i = 1, 2 \ldots h \text{ ,} \tag{6}$$

where $\mathbf{X}_i'$ is the reconstructed image sampled along the direction $\mathbf{P} = [p_1, p_2, \ldots, p_n]$ in the latent space (Fig. 1a), namely the projection-wise disentangling. $\bar{\mathbf{d}} = [\bar{d}_1, \bar{d}_2, \ldots, \bar{d}_n]$ denotes the mean latent representations of datapoints over the testing set, representing a reference point for the sampling. $\mathbf{k}_i$ is a self-defined parameter to control the step of the sampling.

## 3 Experiments

### 3.1 Dataset and Tasks

We applied the proposed method to analyzing how the facial shape is related to patient characteristics and clinical parameters.

The data used in this work is from a multi-ethnic population-based cohort study [24]. We included 9-year-old children that underwent raw 3D facial shape imaging with a 3dMD camera system [15]. We built the raw data into a template-based dataset following Booth's procedures [16]. Additionally, extensive phenotyping was performed regarding gender, BMI, height, ethnicity, low to moderate maternal alcohol exposure (i.e., drinking during pregnancy), maternal age and maternal smoking (during pregnancy). The binary phenotypes were digitalized. Gender: 1 for female and 0 for male; Ethnicity: 1 for Western and 0 for non-Western; Maternal alcohol exposure: 1 for exposed and 0 for non-exposed; Maternal smoking: 1 for exposed and 0 for non-exposed.

The first experiment was designed to extract features only related to the target and independent to defined biases. We therefore investigated the relation between basic characteristics (gender, height and BMI) [25]. When predicting one of the attributes, the other two were considered as biases. The number of labelled samples for this experiment was 4992.

In the second experiment we evaluate the applicability of the method in a clinical practice setting. Low to moderate maternal alcohol consumption during pregnancy could have effects on children's facial shape [17]. We aim to predict if a child was exposed or not, and to explain which part of the face is affected. As suggested by [17], gender, ethnicity, BMI, maternal age and smoking were considered as biases in this task. For this experiment we had access to 1515 labelled samples (760 non-exposed and 755 exposed) and 3496 missing-label samples. The missing-label samples were only involved in the semi-supervised learning settings of our method.

Details about the data characteristic of the two experiments can be found in Fig. A4 and Table A1 in the supplementary files.

### 3.2 Implementation details

For this 3D facial morphology analysis, we implemented the proposed method using a 3D graph convolution network based on Gong's work [18], which was originally designed for unsupervised 3D shape reconstruction. Correlation (Eq. 3) was computed on batch level. For the correlation term (Eq.3), to minimize $|Corr(\mathbf{z_p}, \mathbf{s})|$ we set $\eta$ to 0.5; for the joint loss function (Eq. 5), $L_{rec}$ and $L_{corr}$ were equally weighted ($\lambda = 1$). The training stopped after 600 epochs. For all experiments, the batch size was 64. The number of latent features was 32 for the first experiment, and 64 for the second.

The proposed method was compared with the following state-of-the-art models:

- 3D graph autoencoders (3dAE) [18]: An unsupervised model for 3D shape reconstruction, which serves as a baseline reconstruction method without any restriction on latent features, i.e., does not support fair representation learning.

- VAE-regression (VAE-reg) [19]: A supervised VAE model for interpretable prediction, but unable to handle the bias during training. After training, it is able to reconstruct a list of images by taking as input a list of corresponding $t$.
- VFAE-MI [8]: A supervised VAE-based method for fair representation using MMD loss. Since MMD is not applicable to multiple or continuous biases, we replaced it by MI loss [21], which is considered to be equivalent or better than the MMD loss in some applications [11].
- BR-Net [7]: A supervised adversarial-training-based method, which uses statistical correlation metric as the adversarial loss rather than the commonly used cross-entropy or MSE losses. The original method was designed to handle single bias only. As the authors suggested [20], we adapted the method to handle multiple biases by adding one more BP network for each additional bias.

### 3.3 Evaluation Metrics

In this paper, a fair prediction means the prediction $\hat{t}$ is unbiased by $s$; Fairness in disentanglement is to disentangle facial features related to the target but not confounded by bias.

Prediction results were evaluated on two aspects: the prediction accuracy and fairness. For binary prediction, the accuracy was quantified by the area under the receiver operating characteristic curve (AUC); for prediction of continuous variables, the root mean square error (R-MSE) was measured. The fairness of the prediction was quantified by $|Corr(\hat{t}, \mathbf{s_i})|$ (simplified as $|Corr(\mathbf{s_i})|$) in the range of [0,1], which measures to what extent the prediction is biased by the bias. In Table 1-2 we added '+' or '-' for $|Corr(\mathbf{s_i})|$, in order to tell if there is an overestimation (+) or underestimation (-) in the prediction when given a larger $s_i$. For autoencoder-based methods, the reconstruction error (Rec error) was quantified by mean $L_1$ distance. All results were based on 5-fold cross-validation.

Since only VAE-reg [19] is interpretable, the proposed method is compared with it in terms of interpretability. To qualitatively evaluate the interpretation performance, we provide ten frames of faces reconstructed by Eq. 6 (Fig. A2), and the corresponding difference heatmap between the first and the last frame. For each task, $k_i$ in Eq. 6 was adjusted to control the range of $\hat{t}_i$ to be the same as that of VAE-reg [19].

## 4 Results

### 4.1 Phenotype prediction for gender, BMI, and height

**Fair Prediction.** Compared to other fair methods, our method overall obtained the best fair prediction accuracy (R-MSE and AUC), while controlling the biases information at the lowest level ($|Corr(\mathbf{s_i})|$) (Table 1). Compared with VFAE-MI, the proposed method showed much better reconstruction quality, with similar performance to 3dAE. In addition, we observed a more robust training procedure of the proposed method than the compared VAE- and adversarial-based methods (Fig. A1 in supplementary).

**Table 1.** Fair prediction of gender, BMI, and height. Column X shows the correlation between the ground truth **t** and the biases $s_i$. '+' and '–' means there was respectively an overestimation and underestimation in the prediction when given a larger $s_i$. The best result for each row among 'Methods with fairness' is in bold.

| | Methods without fairness | | | Methods with fairness | | |
|---|---|---|---|---|---|---|
| | X | 3dAE | VAE-reg | VFAE-MI | BR-Net | Ours |
| *Gender prediction* | | | | | | |
| AUC ↑ | 1 | - | 0.866 | 0.807 | 0.802 | **0.840** |
| \|Corr (BMI)\| ↓ | +0.035 | - | -0.061 | -0.037 | -0.035 | **-0.035** |
| \|Corr (height)\| ↓ | -0.033 | - | -0.133 | -0.072 | -0.073 | **-0.047** |
| Rec error ↓ | - | 0.273 | 0.275 | 0.671 | - | **0.295** |
| *BMI prediction* | | | | | | |
| R-MSE ↓ | 0 | - | 1.786 | 2.706 | 2.504 | **2.373** |
| \|Corr (height)\| ↓ | +0.243 | - | +0.338 | +0.056 | +0.047 | **+0.023** |
| \|Corr (gender)\| ↓ | +0.035 | - | +0.027 | +0.034 | +0.021 | **+0.013** |
| Rec error ↓ | - | 0.273 | 0.276 | 0.681 | - | **0.278** |
| *Height prediction* | | | | | | |
| R-MSE (cm) ↓ | 0 | - | 5.303 | 6.617 | 6.480 | **6.222** |
| \|Corr (BMI)\| ↓ | +0.243 | - | +0.424 | +0.030 | +0.026 | **+0.013** |
| \|Corr (gender)\| ↓ | -0.033 | - | -0.201 | -0.049 | -0.038 | **-0.028** |
| Rec error ↓ | - | 0.273 | 0.277 | 0.738 | - | **0.278** |

**Interpretation.** Fig. 2 provides visualizations of the facial features that are used by the methods for prediction tasks. The baseline model (VAE-reg) [19] captured all features to boost the prediction, whereas our model captured the features that only related to the target and independent to biases. For gender prediction, our result is similar to that of VAE-reg because gender is nearly unbiased by height and BMI in the dataset. Since BMI and height are positively correlated in our dataset, the similar heatmaps for the BMI and height prediction of the VAE-reg indicate that it captured common facial features for the two tasks, and thus failed to disentangle the confounding bias. In contrast, our model learned a target-specific representation, showing a strong correlation to the target and without being confounded by other (bias) attributes.

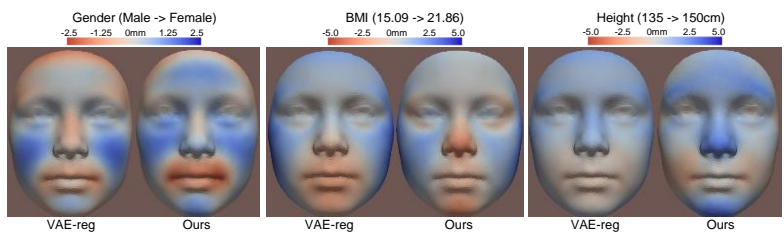

**Fig. 2.** Interpretation of facial features extracted for prediction. VAE-reg: results of VAE-reg [19]; Ours: results of the proposed method. Red and blue areas refer to inner and outer facial changes towards the geometric center of the 3D face, respectively.

### 4.2 Prediction on maternal alcohol consumption during pregnancy

**Fair Prediction.** In the second experiment, our method achieved a similar prediction accuracy as the other fair methods while controlling the bias information at the lowest level (Table 2). Compared with the results of 'Ours', the results of 'Ours-SSL' showed that the semi-supervised strategy further improved the prediction accuracy and reconstruction quality by additionally including missing-label training data, when controlling the bias information at a similar level.

**Table 2.** Fair prediction results. Ours-SSL refers to the semi-supervised settings of our method.

|  | Methods without fairness | | | Methods with fairness | | | |
|---|---|---|---|---|---|---|---|
|  | X | 3dAE | VAE-reg | VFAE-MI | BR-Net | Ours | Ours-SSL |
| AUC ↑ | 1 | - | 0.768 | 0.572 | 0.563 | 0.579 | **0.587** |
| \|Corr (ethnicity)\| ↓ | +0.479 | - | +0.430 | +0.076 | **+0.024** | +0.040 | +0.037 |
| \|Corr (maternal smoking)\| ↓ | +0.288 | - | +0.125 | +0.050 | +0.044 | **+0.032** | +0.039 |
| \|Corr (maternal age)\| ↓ | +0.407 | - | +0.257 | +0.054 | +0.044 | **+0.031** | +0.034 |
| \|Corr (BMI)\|↓ | -0.318 | - | -0.418 | -0.079 | -0.055 | -0.030 | **-0.019** |
| \|Corr (gender) \|↓ | -0.020 | - | +0.078 | +0.070 | +0.069 | +0.044 | **+0.036** |
| Rec error ↓ | - | 0.276 | 0.331 | 0.656 | - | 0.344 | **0.316** |

**Interpretation.** We compared the interpretation results of the VAE-reg and our methods (Fig. 3), and further explained the gap between the baseline and our results, by visualizing our results with gradually increased fairness (Fig. 4): from left to right, in the first figure $L_{corr}$ corrected for no bias in the training; in the last figure $L_{corr}$ corrected for all bias in the training (Eq. 4). Our results suggest low to moderate maternal alcohol exposure during pregnancy affected children's facial shape. Affected regions were shown by our methods in Fig. 3, which is consistent with existing findings [17].

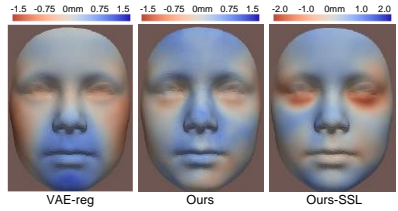

**Fig. 3.** Heatmaps show how the facial shape changes from non-exposed to exposed.

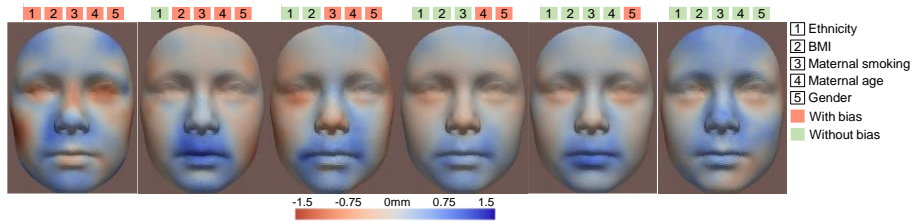

**Fig. 4.** Explanation about how the target features were disentangled from biases.

## 5 Discussion and Conclusion

In this paper, we proposed a projection-wise disentangling method and applied it to 3D facial shape analysis. For evaluated tasks (BMI, height, gender and maternal alcohol exposure prediction), we achieved the best prediction accuracy while controlling the bias information at the lowest level. Also, we provide a solution to interpret prediction results, which improved mechanistic understanding of the 3D facial shape. In addition, given the shortage of labelled data in many domains, we expect that the proposed method with its semi-supervised extension can serve as an important tool to fully exploit available data for fair representation learning.

Beyond the presented application, the proposed method is widely applicable to prediction tasks, especially to clinical analysis where confounding bias is a common challenge. For future work, we plan to disentangle aging effects from pathology of neurodegenerative diseases.

The improvement of our method mainly comes from the projection strategy. Previous methods learn fair features $\mathbf{Z}$ by forcing $MI(\mathbf{Z}, s) \rightarrow 0$, which encourages a global independence between $\mathbf{Z}$ and s, i.e., **any linear or non-linear combination of Z** contains no information of $s$. This strong restriction leads to a decrease in diversity of learned features (Fig. A3), resulting in huge reconstruction error in VFAE-MI (Table 1-2). This loss of diversity also explains why the prediction accuracy of baselines were limited although using all features in $\mathbf{Z}$ for prediction. In contrast, our strategy can be viewed as a conditional independence between $\mathbf{Z}$ and $s$, i.e., **only a linear combination of Z** (the projection $z_p$) being independent to $s$. This strategy allows that most of the information can be kept in the latent space, and thus minimizing the conflicts between reconstruction quality and fairness in auto-encoder models. This is crucial especially when the biases contain much more information of the input than the target does, e.g., the second experiment.

## Appendix

| **Algorithm 1**: proposed semi-supervised fair representation learning |
| --- |

**Require:** $X^u$ (unlabelled data), $X^l$ (labelled data), $N_{epoch}$, $N_{batch\_size}$

**Parameters:** $\theta_{enc}$, $\theta_{dec}$, $\theta_{pe}$

1.  $N_{sample} = \min \{len(X^u), len(X^l)\}$
2.  i = 0
3.  **while** i < $N_{epoch}$, **do**:
4.    randomly sample $X^{us}$, $X^{ls}$ from $X^u$, $X^l$, respectively. $len(X^{us}) = len(X^{ls}) = N_{sample}$
5.    j = 0

6.     **while** $j < N_{sample}$, **do**:
7.         Acquire *unlabelled* datapoints $\mathbf{X}^{ut}$ from $\mathbf{X}^{us}$. $len(\mathbf{X}^{ut}) = N_{batch_{size}}/2$
8.         update $\theta_{enc}$ and $\theta_{dec}$ by $L_{rec}(\mathbf{X}^{ut})$
9.         Acquire *labelled* datapoints $\mathbf{X}^{lt}$ from $\mathbf{X}^{ls}$. $len(\mathbf{X}^{lt}) = N_{batch\_size}/2$
10.       update $\theta_{enc}$, $\theta_{dec}$ and $\theta_{pe}$ by $L_{joint}(\mathbf{X}^{lt})$
11.       $j = j + N_{batch\_size}/2$
12.     **end while**
13.     i = i + 1
14. **end while**
15. **return** $\theta_{enc}$, $\theta_{dec}$, $\theta_{pe}$

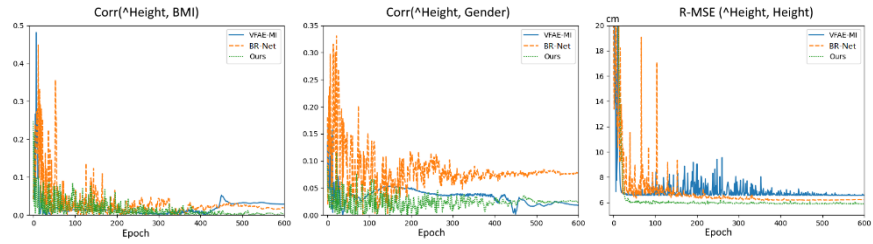

Fig. A1: Validation curves of the methods for the height prediction task (Table 1).

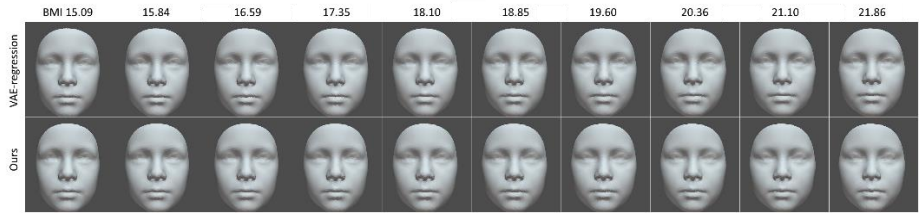

Fig. A2: 3D faces reconstructed by VAE-regression (1st row), and by the proposed projection-wise disentangling (2nd row) for the BMI prediction task.

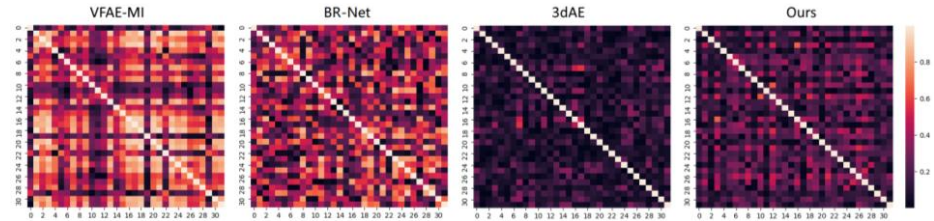

Fig. A3: Correlation matrix of the learned latent features by VFAE-MI, BR-Net, 3dAE, and our method for the height prediction task. Higher correlations indicate lower feature diversity.

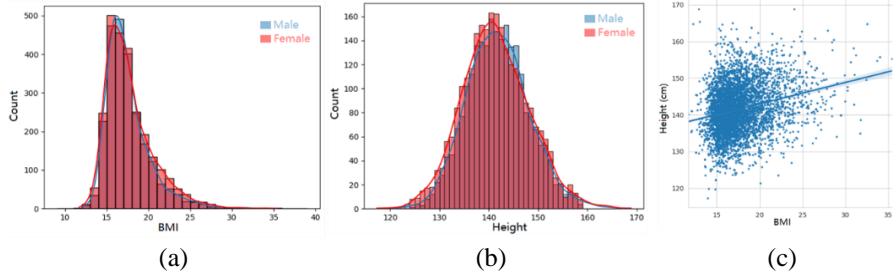

(a)  (b)  (c)

Fig. A4: Data characteristic for the first experiments (N=4992). a) Histogram of BMI, for male and female, respectively; b) Histogram of height, for male and female, respectively; c) Joint distribution of BMI and height. A positive correlation between BMI and height can be observed.

**Table A1.** Data characteristic of children and their mothers included in the second experiment (for labelled data N=1515). Maternal smoking, maternal age, child's BMI and especially ethnicity showed imbalanced distribution between the non-exposed and exposed groups.

| Characteristic | Non-exposed (control: N=760) | Exposed (case: N=755) |
|---|---|---|
| Child's ethnicity, No. (%) | | |
| Western | 367 (48.3) | 670 (88.7) |
| Non-western | 393 (51.7) | 85 (11.3) |
| Child's gender, No. (%) | | |
| Male | 357 (47.0) | 370 (49.0) |
| Female | 403 (53.0) | 385 (51.0) |
| Child's BMI, mean (SD) | 18.6 (3.2) | 16.8 (2.0) |
| Maternal smoking, No. (%) | | |
| Yes | 204 (26.8) | 417 (55.2) |
| No | 556 (73.2) | 338 (44.8) |
| Maternal age, mean (SD) | 28.2 (5.0) | 32.1 (3.9) |

**Implementation details for baseline models:**

For VAE-reg, VFAE-MI, BR-Net, we adopted the default configurations from their official implementations (see links below). For this 3D facial shape analysis, we replace the original convolutional networks by the graph convolutional networks (3dAE, see links below). Four layers of graph convolutional networks were implemented for feature extraction, and another four layers for decoder in VAE-based models. These settings guaranteed both baselines and our method had the same network framework for feature extraction (and for decoder if applicable). Any changes beyond above-mentioned settings were described in section 3.2. For both baselines and our method, the number of latent features was 32 for the first experiment, and 64 for the second.

We will release the code for all methods in:
https://github.com/tsingmessage/projection_wise_disentangling_FRL

| | |
|---|---|
| VAE-reg: | https://github.com/QingyuZhao/VAE-for-Regression |
| VFAE-MI: | https://github.com/dendisuhubdy/vfae |
| BR-Net: | https://github.com/QingyuZhao/BR-Net/ |
| 3dAE: | https://github.com/sw-gong/spiralnet_plus |